\documentclass[letterpaper]{article}
\usepackage[preprint]{aaai2027}
\usepackage[hyphens]{url}
\usepackage{graphicx}
\usepackage{natbib}
\usepackage{booktabs}
\DeclareTextFontCommand{\texttt}{\sffamily\upshape}

\title{Monitor-Validity Failures: LLM Safety Evidence Paths\\
at Evolving Framework Boundaries}
\author{Junchi Liao}
\affiliations{}

\begin{document}
\maketitle

\begin{abstract}
Modern LLM applications rely on runtime monitors that inspect tool calls, traces, streams, detector outputs, and verifier inputs.
This creates an interface-reliability problem: a framework path may remain operational while policy-relevant evidence never reaches its decision point.
In one representative case, a regex detector returned a positive match, but the framework stranded the result before the block decision.
This differs from jailbreaks and detector errors because the evidence is already recognized yet disconnected from enforcement.
We study this repair-boundary signature with canary contracts: parent/fixed confirmation units that place a known payload only on the changed evidence surface and record monitor visibility together with the downstream decision for action-boundary cases.
Applying the protocol to public framework repairs, we confirm 18 evidence-path failures across 11 LLM or agent safety project groups and four implementation ecosystems.
Nine are established-boundary repairs, eight are new-surface policy-coverage gaps, and one is a boundary-only evidence case.
Fourteen reach E1 enforcement boundaries: seven native action outcomes, six configured enforcement boundaries, and one synthetic-consumer consequence.
The remaining cases affect verifier, safety-prompt, or observability channels.
The cases form carrier, adapter, authority, and surface families and show how ordinary application behavior can remain intact while a safety-specific evidence path misses its decision.
Exact revisions, deterministic outcomes, negative controls, and a frozen prospective audit document the confirmation rule.
Monitor assumptions therefore need regression tests at the schemas, streams, tools, and adapters that frameworks evolve.
\end{abstract}

\section{Introduction}

A guardrail in a modern LLM application is rarely just a prompt wrapper.
Before it blocks, routes, redacts, or reports, it may read request objects, tool-call arguments, stream events, generated files, detector outputs, or verifier inputs.
Safety therefore depends on structured evidence that sits between the prompt and the final response \citep{inan2023llamaguard,han2024wildguard,wang2025agentspec}.

A reproduced NeMo Guardrails repair shows the problem in miniature.
The regex detector returned a structured positive result with \texttt{is\_match=true}.
The parent framework still allowed the output, because that positive result was never mapped to the block decision.
The fixed revision added the missing output mapping, and the same canary was blocked.
Nothing here calls for a stronger jailbreak prompt or a better regex.
The detector had seen the evidence; the framework had dropped it before enforcement.

The NeMo case is only one instance of a broader software problem.
Request schemas change, tool-call representations expand, multimodal fields appear, streaming surfaces are added, and policy adapters are refactored.
Agent and tool-use systems make these surfaces part of ordinary execution \citep{yao2023react,schick2023toolformer,qin2023toolllm}.
A monitor can stay configured and stay in the loop while the evidence it relies on moves to a new field, loses metadata, reaches a different adapter, or is materialized only after the safety check.
We use monitor-validity failure for the observable parent/fixed repair contrast and reserve monitor-validity drift for its longitudinal working-to-failing subtype.

That failure is easy to miss when the problem is framed as another jailbreak or guardrail benchmark \citep{wei2023jailbroken,mazeika2024harmbench,zhan2024injecagent,debenedetti2024agentdojo,andriushchenko2025agentharm}.
Those evaluations ask whether unsafe behavior can be elicited or recognized through a specified interface.
Our question is different: once evidence is recognized, does the exercised framework path carry it to the policy boundary?
A strong detector, policy DSL, verifier, or LLM-as-judge component cannot protect a reproduced boundary when the parent path does not deliver the evidence or positive result it was designed to use.

Source diffs and issue reports are not enough to confirm this miss.
The parent revision must already have an intended boundary, a deterministic payload must cross the changed surface, and the parent and fixed revisions must differ in visibility or decision behavior.
We use canary contracts as the confirmation unit.
Each contract places a known policy-relevant payload only on the changed evidence surface and records monitor visibility together with downstream decision behavior when the case claims an action boundary.

\begin{figure*}[t]
\centering
\includegraphics[width=\textwidth]{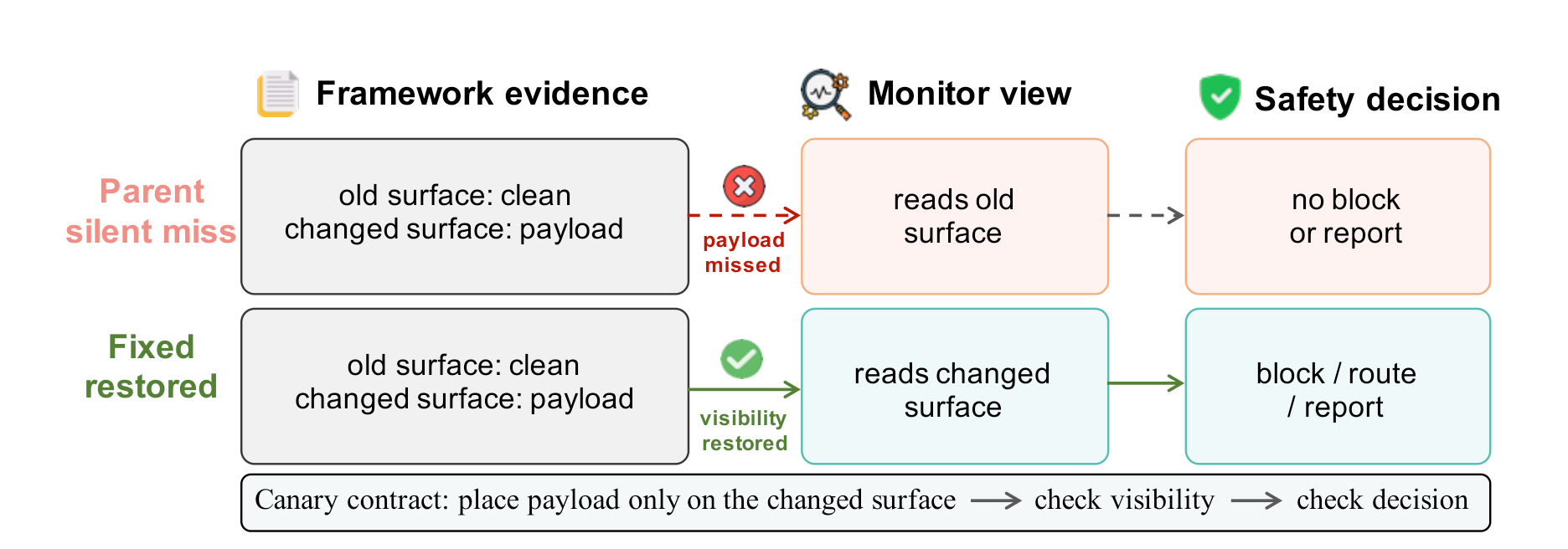}
\caption{Monitor-validity failure at a repair boundary.
The parent monitor still runs but reads an incomplete evidence surface; the repaired child reconnects the same canary payload to monitor visibility and decision.}
\label{fig:overview}
\end{figure*}

Using canary contracts, we study 18 confirmed repair-boundary cases across 11 framework project groups and four implementation ecosystems.
We keep the evidence strata explicit: 9 are established-boundary repair cases, eight are new-surface validity gaps, and one remains boundary-only evidence.
We also keep reproduction strength separate from impact: 14 cases reach E1 enforcement boundaries, comprising seven native action outcomes, six configured enforcement boundaries, and one synthetic-consumer consequence, while the remaining cases affect verifier, safety-prompt, or observability channels.
We organize the cases into four dependency families, depending on where the hidden assumption broke: the payload carrier, the detector-to-action adapter, the authority among competing channels, or the monitored surface itself.

This framing also fixes the unit of evaluation.
We do not ask whether a model is unsafe in general.
We ask whether a specific safety boundary still receives the evidence it was built to use.
That distinction matters for regression testing: the failing object is often a schema, adapter, stream, or callback rather than a prompt.

This paper makes three contributions:
\begin{itemize}
\item We provide a case-level corpus of 18 repair-boundary confirmations across 11 framework project groups, with exact revisions, deterministic canaries, and explicit evidence, impact, subtype, and oracle records.
\item We organize the cases into four structural failure families that explain how hidden evidence-to-decision assumptions break under framework evolution.
\item We use canary contracts as safety-specific differential regression-test templates, and we report a five-case runnable diagnostic comparison of full canary contracts with partial checks.
\end{itemize}

\section{Background and Problem Setting}

The examples above share a simple shape: safety logic reads a framework representation, not the world directly.
This section fixes the vocabulary for the runtime objects, decisions, and change patterns used in the study.

\subsection{Runtime Safety Monitors}

We treat a runtime monitor as any component that observes framework state and can affect a safety-relevant decision.
The state may be a message, tool call, stream delta, trace metadata field, detector result, provider metadata object, or artifact path.
The decision may block a request, redact output, interrupt a tool call, route a filter result, cap automatic tool execution, or produce a verifier finding.
The important dependency is structural: the policy boundary only works if it still sees the runtime representation that carries policy-relevant state.

\subsection{Monitor-Validity Failures at Repair Boundaries}

Monitor validity is relative to the representation and routing path that a policy boundary actually uses.
We empirically study repair-boundary failures: in a reproduced parent revision, a policy-relevant canary is absent from the monitor-visible evidence or fails to reach the expected decision; in the fixed revision, the same canary reaches the corresponding visibility or decision oracle.
This two-point contrast establishes a failing parent and repaired child, not an earlier working revision.

For example, a request guardrail may inspect user text in a \texttt{messages} field while a supported request path carries the policy-relevant payload through \texttt{input}.
The application path can complete normally even though the guardrail receives an empty or stale view.
Other cases preserve the evidence but strand a detector result, skip a configured hook, or select the wrong authority before the decision boundary.

The failure is silent in this mechanical sense.
The changed framework path need not crash, and the monitor need not disappear from configuration.
Instead, policy-relevant state crosses a dependency boundary that the monitor treats as complete.
For a given monitor, only part of the framework state is visible: message fields, tool-call objects, stream chunks, detector outputs, callback arguments, or verifier inputs.
When several fields describe the same event, the monitor also depends on which field the framework treats as controlling, such as free-text content or structured tool-call metadata.

We reserve monitor-validity drift for a working-to-failing-to-repaired temporal subtype; none of our 18 cases establishes that full interval.
Our cases establish only failing-parent/repaired-child repair contrasts.
A canary contract tests that contrast by placing the same payload on the exercised surface and checking visibility and, where applicable, decision behavior.

A confirmed repair-boundary case compares a parent revision and a fixed revision around a public repair.
The relevant parent-side framework path must remain operational, but the payload must be absent from monitor-visible evidence or fail to reach the decision.
The fixed side must connect the same payload to the corresponding visibility or decision oracle.
We do not require that the parent revision introduced the failing condition; it is the nearest reproduced failing side of a public repair boundary.
A canary contract packages this check into a small before/after assertion over visibility and, whenever the case reaches an action boundary, block, route, redact, interrupt, or finding behavior.

\subsection{Scope}

The study is targeted and deterministic rather than prevalence-oriented.
The empirical contribution is a documented set of 18 confirmed repair-boundary cases where a safety monitor and an evolving framework disagree about the evidence that reaches a policy boundary.
Cases were selected for feasibility and determinism, and the corpus reports confirmation, reproduction strength, impact, and subtype as separate axes.
Detailed provenance and reproduction protocols are provided with the case records and artifact.

\section{Canary Contracts: A Method for Confirming Silent Boundary Failures}

Suspicious patches are easy to find; confirmed failures are not.
For each candidate, we need to show that the parent version already had a relevant safety boundary, that a policy-relevant payload crossed the changed surface, and that the fixed version reconnects that payload to visibility or action.
We use a canary contract for this check: one payload, one repaired boundary, and two observations about the same execution.

The contract isolates interface reliability from prompt or model quality.
It places a known payload exactly on the changed carrier, adapter result, stream, or authority channel, then compares what the parent and fixed versions expose and which decisions they make.
In our cases, that payload is a small deterministic marker chosen for the boundary: a denied tool name, a regex-matching token, a synthetic credential, a tool argument, or a vulnerable code snippet.
Its important property is placement, not linguistic difficulty.

\subsection{The Confirmation Problem}

Silent failures leave little obvious evidence in ordinary execution.
The application may return normally, the monitor may still be configured, and logs may show that some guardrail code ran.
A source diff or issue title is therefore not enough.
The same patch could repair a real monitor miss, remove an old boundary, add a new feature, or refactor code without changing behavior.
We only count the first case, so the protocol has to isolate the evidence-to-decision path: the parent must miss the payload, and the fixed version must restore visibility or enforcement.

\subsection{Canary Contract Definition}

In practice, a contract records five things: the repaired version pair, the intended monitor boundary, the canary placement, a visibility check, and a decision check.
The canary payload is a tracer for the evidence path, not an attack prompt: the test asks whether safety-relevant evidence still reaches the policy boundary.
The version pair fixes the software change under study.
The boundary explains why the parent should already have observed or enforced the payload.
The placement keeps the payload off legacy surfaces, so a passing fixed result cannot be explained by an unrelated detector path.
The two checks separate seeing the evidence from acting on it.

\begin{table}[t]
\centering
\caption{Canary-contract fields used for confirmation.}
\label{tab:contract}
\small
\begin{tabular}{ll}
\toprule
Contract field & Role in confirmation \\
\midrule
Version pair & fix the parent/repaired revisions \\
Boundary & exclude new features or removed policies \\
Canary & place payload only on the changed surface \\
Visibility oracle & check monitor visibility of the payload \\
Decision oracle & check whether the intended action fires \\
\bottomrule
\end{tabular}
\end{table}

Visibility and decision are separate because they fail in different places.
A detector can return a positive match while an adapter lets the output through; a blocking path can work for chat messages while a newer request shape remains invisible.
In the regex example, the visibility check asks whether the detector matched, while the decision check asks whether that match blocked the output.
A case is confirmed only when the parent shows the miss and the fixed version shows the repair.
Candidates that crash, fail closed, or depend on model sampling are recorded separately.

Viewed this way, a canary contract is a regression test with a safety-specific oracle.
Ordinary regression tests ask whether functional behavior is preserved.
This test asks whether a policy intent is still connected to the evidence surface that changed.
The useful part is the placement: put the payload only on the new carrier, stream, adapter result, or authority channel, then record monitor visibility and, at action boundaries, the safety action.

\section{Case Selection and Corpus Construction}

We built the corpus from public repair references, issue trails, commit scans, and project-specific triage.
A reconstructed denominator pass contributed 195 high-signal public PR and issue references using monitor and failure-indicator queries; every candidate source was subjected to the same confirmation rule.
Appendix B records the query flow, candidate-entry routes, and exclusions (Table 9).

Promising references moved forward only after a deterministic parent/fixed contrast could be reproduced.
Otherwise they stayed as notes, boundary cases, or negative controls.
A case is counted only when all six conditions hold: identifiable parent and fixed revisions; evidence of an intended safety contract for the affected or equivalent surface; a deterministic payload placed only on the changed surface; execution of the parent-side monitor path; a parent miss in visibility or decision; and a fixed-side repair of the same visibility or decision.
We treated a boundary as intended only when supported by configuration semantics, prior behavior on an equivalent surface, an upstream or fixed regression test, documentation, or issue/PR discussion.
Loud crashes, unmerged PRs, ordinary feature additions, generic auth or rate-limit bugs, and fail-closed parents were excluded.
Appendix D records the intent-evidence type for each confirmed case (Table 13).

Subtype assignment is separate from confirmation.
Established-boundary cases have an intended affected boundary on the parent side; new-surface cases connect an existing policy expectation to a newly supported representation.
The boundary-only row preserves a deterministic repair contrast while keeping its temporal subtype separate from the established and new-surface groups.

We record the primary source of intent evidence for every case: an existing test or fixture for eight cases, issue or pull-request discussion for five, configuration or documented semantics for four, and equivalent-surface behavior for one.
Appendix D preserves the richer case-level source descriptions.

For each promoted candidate, we wrote or backported a canary that places the payload on the changed surface and checks visibility and decision whenever the case claims an enforcement boundary.
The confirmed corpus contains 18 cases across 11 framework projects and four implementation ecosystems.
Project counts use framework-level groups: for example, Semantic Kernel Python and .NET are one project group, and the LiteLLM/Dynamiq stream case is counted under the LiteLLM boundary.
Together, these procedures produce a confirmed cross-framework corpus with a uniform parent/fixed behavioral oracle.

\subsection{Evidence and Impact Tiers}

Table~\ref{tab:summary} summarizes the corpus.
We report two axes: reproduction strength and demonstrated impact.
Reproduction strength records how directly the canary reaches the changed monitor boundary: Tier A uses framework-level or module-level execution; Tier B isolates the changed logic with stubs around unrelated dependencies.
Impact records the strongest safety boundary shown by the canary: E1 for enforcement, blocking, interruption, rewrite, filter, or routing decisions; E2 for verifier or safety-prompt evidence decisions; E3 for evidence visible only in an observability channel.

The table also reports known-pattern fixture patterns.
These small local fixtures check whether a metadata emitter can express the same visibility and decision contract shape across all four observed pattern classes.

The rows serve different parts of the empirical claim.
The 18 confirmed cases support the before/after behavior claims; the project, ecosystem, tier, and impact rows describe the breadth and boundary of those cases.
The fixture and denominator rows are audit records: they check contract expression and candidate selection, but they are not additional failures or a prevalence sample.

\begin{table}[t]
\centering
\caption{Summary of the confirmed corpus and validation records.}
\label{tab:summary}
\small
\begin{tabular}{lr}
\toprule
Study item & Count \\
\midrule
Confirmed before/after cases & 18 \\
Established / new-surface / boundary-only & 9 / 8 / 1 \\
Framework projects & 11 \\
Implementation ecosystems & 4 \\
Tier A / Tier B confirmations & 17 / 1 \\
E1 / E2 / E3 impact boundaries & 14 / 3 / 1 \\
Known-pattern fixture families & 14 \\
Manually labeled denominator refs & 195 \\
\bottomrule
\end{tabular}
\end{table}

These axes distinguish execution depth from demonstrated impact.
Tier A executes framework or module monitor code, while Tier B isolates changed logic around unrelated dependencies.
E1 cases exercise action boundaries and E2 cases exercise verifier or safety-prompt evidence.
All 18 cases meet the canary-contract confirmation rule.

\section{Experiments}

The confirmation protocol gives us a concrete unit of evidence: a parent miss and fixed repair for a specific monitor boundary.
The experiments ask four questions in the order a skeptical reader needs them answered: what breaks, whether confirmed repair-boundary cases span frameworks, how far misses reach into safety decisions, and what controls bound the method claims.
The confirmed case remains the unit throughout; project-level counts measure only cross-framework breadth.

\subsection{RQ1: What Assumptions Break Under Framework Evolution?}

We first characterize what breaks when these repair-boundary evidence-path failures occur.
We organize the 18 confirmed repair-boundary cases into four structural failure families, each corresponding to a different hidden assumption about the path from evidence to decision.
The family label marks the first broken dependency on the exercised evidence-to-decision path.

\begin{figure}[t]
\centering
\includegraphics[width=\linewidth]{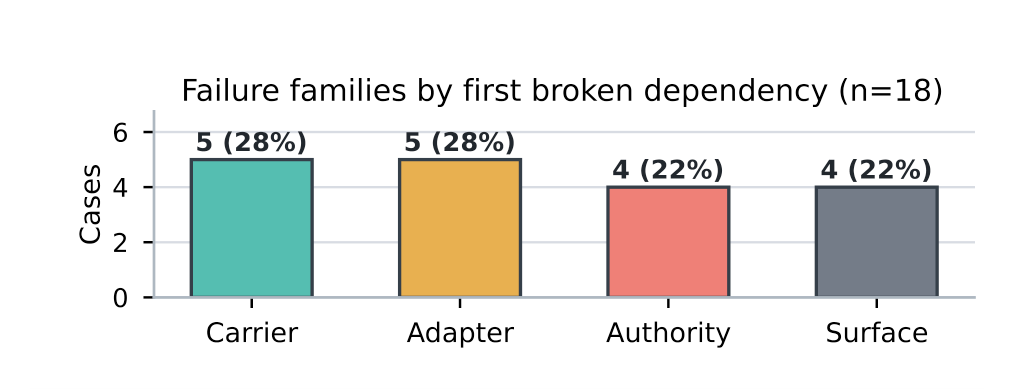}
\caption{Distribution of the 18 cases by first broken evidence-to-decision dependency; percentages are descriptive corpus shares.}
\label{fig:families}
\end{figure}

\begin{table}[t]
\centering
\caption{Failure families and the canary focus used to test each hidden assumption.}
\label{tab:families}
\small
\begin{tabular}{lll}
\toprule
Family & Hidden assumption & Canary focus \\
\midrule
Carrier & payload location & new carrier only \\
Adapter & result-to-action map & positive result \\
Authority & channel priority & disagreement \\
Surface & materialization order & new surface only \\
\bottomrule
\end{tabular}
\end{table}

Figure~\ref{fig:families} gives the descriptive split in this targeted corpus.
We do not use these counts as prevalence evidence.
The useful distinction is where the hidden assumption breaks and where the canary must be placed.
A carrier canary tests whether the payload remains attached to the object the monitor reads; an adapter canary tests whether a positive monitor result reaches an action; an authority canary forces two channels to disagree; and a surface canary places the payload only on the newly materialized request or stream surface.

The established-boundary subset sharpens the repair claim.
Restricted to the 9 cases where an existing intended monitor boundary misses evidence or action on the reproduced parent side, the corpus has Adapter, Authority, and Carrier failures (4, 3, and 2 cases respectively), all at E1 action boundaries, across eight project groups.
The Surface cases in this corpus are new-surface validity gaps, so we treat them as related repair-boundary failures rather than temporal regressions.

\begin{figure}[t]
\centering
\includegraphics[width=\linewidth]{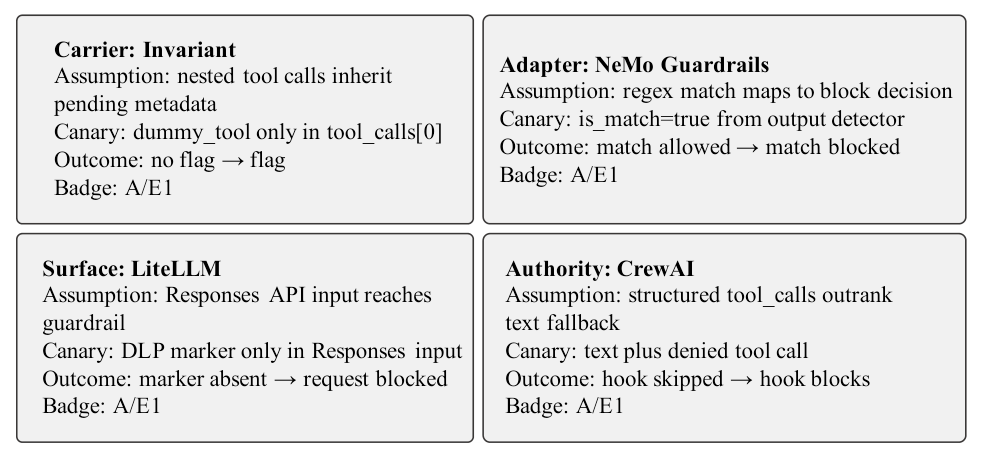}
\caption{Representative canary contracts from the confirmed corpus, one per failure family.
The full 18-case audit matrix appears in Appendix D.}
\label{fig:cards}
\end{figure}

Figure~\ref{fig:cards} shows one representative contract per family.
Each card records the assumption being tested, the targeted canary placement, the parent-to-fixed outcome, and the reproduction/impact badge.

Carrier failures are silent because the monitored object still exists, but the policy-relevant payload has moved to a different carrier.
Invariant is the canonical example: the policy analyzer matched a nested tool call, but pending-trace metadata was attached only to the top-level message, so the violation was filtered as not pending.
The correct canary therefore puts the forbidden payload only on the nested carrier and checks both metadata visibility and policy decision.

Adapter failures preserve detection but break enforcement.
In the NeMo regex case, the detector returned a structured positive result, but the parent action lacked the output mapping that turns that result into a block.
The monitor path therefore looked alive: the detector ran and returned a result.
The missing contract was the detector-output-to-action adapter, so the canary must force a positive result and assert that it reaches the blocking decision.

Authority failures arise when a framework exposes multiple plausible channels for the same event.
A tool call may appear in structured metadata while a text field remains benign, or a public agent identity may be replaced by an internal execution clone.
These cases are silent because downstream policy logic reads a valid-looking but lower-authority representation.
Their canaries make two channels disagree and assert that the monitor uses the channel intended by the policy boundary.

Surface failures occur when frameworks add or reorder request, stream, or multimodal surfaces.
A monitor that still inspects text messages can miss image parts, Responses API inputs, or live stream chunks if those surfaces are materialized after inspection.
The surface canary keeps the payload absent from the legacy surface and places it only on the newly supported surface.
This makes the failure a structural coverage miss rather than a detector-quality miss.

Each family requires a different regression contract.
A generic jailbreak prompt may put the same payload in every field and make both parent and fixed versions pass.
These canaries are small because placement, not prompt strength, is the test.

\subsection{RQ2: Do Confirmed Cases Appear Across Frameworks?}

The taxonomy above would be less important if it described a pathology in one library.
We next ask whether the confirmed repair-boundary cases appear across projects, monitor forms, and implementation ecosystems.

We use three breadth checks.
The figure summarizes project and monitor boundaries, the ecosystem matrix checks whether observed families appear outside one implementation language, and the project table checks whether the corpus is dominated by a single large repository.
Together, the three views distinguish structural recurrence from concentration in one repository or language.

\begin{figure}[t]
\centering
\includegraphics[width=\linewidth]{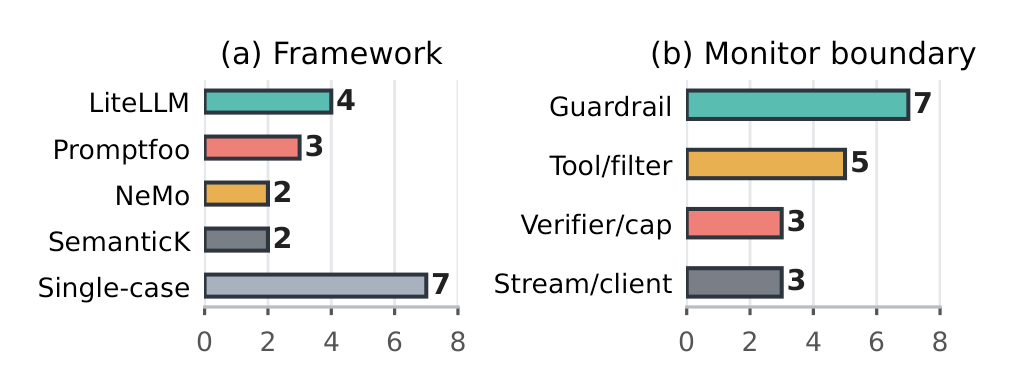}
\caption{Descriptive composition of the 18 confirmed repair-boundary cases; these convenience-corpus counts are not prevalence estimates.}
\label{fig:breadth}
\end{figure}

\begin{table*}[t]
\centering
\caption{Impact-boundary breakdown for confirmed misses.
E1 is split by enforcement form; its 14 cases comprise seven native actions, six configured boundaries, and one synthetic-consumer consequence.}
\label{tab:impact}
\small
\begin{tabular}{lllp{0.42\textwidth}l}
\toprule
Tier & Boundary type & Cases & Broken safety effect & Example \\
\midrule
E1 & block / filter / rewrite & 5 & unsafe content passes in parent; repair blocks or rewrites it & NeMo regex \\
E1 & interrupt / HITL route & 3 & tool execution or routing bypasses the intended stop point & Haystack \\
E1 & hook / cap / policy state & 6 & configured enforcement hook, cap, or rule update does not fire & CrewAI \\
E2 & verifier / safety prompt & 3 & evidence is missing from evaluator or safety-prompt input & Promptfoo \\
E3 & observability channel & 1 & structured evidence is hidden from a policy-relevant consumer & AutoGen \\
\bottomrule
\end{tabular}
\end{table*}

\begin{table}[t]
\centering
\caption{Breadth of failure families across implementation ecosystems.
Language columns count confirmed repair-boundary cases; Projects counts distinct framework project groups touched by each family.}
\label{tab:lang}
\small
\begin{tabular}{lrrrrr}
\toprule
Family & Py & TS & Java & C\# & Projects \\
\midrule
Carrier & 3 & 2 & 0 & 0 & 4 \\
Adapter & 4 & 0 & 0 & 1 & 3 \\
Authority & 2 & 2 & 0 & 0 & 4 \\
Surface & 3 & 0 & 1 & 0 & 4 \\
\bottomrule
\end{tabular}
\end{table}

\begin{table}[t]
\centering
\caption{Descriptive confirmed-case counts by project family.}
\label{tab:projects}
\small
\begin{tabular}{lll}
\toprule
Project group & Cases & Boundary \\
\midrule
LiteLLM & 4 & proxy, stream, policy \\
Promptfoo & 3 & verifier, tool-use \\
NeMo Guardrails & 2 & output, multimodal \\
Semantic Kernel & 2 & termination, filters \\
Single-case projects & 7 & analyzer, client, hook \\
\bottomrule
\end{tabular}
\end{table}

Figure~\ref{fig:breadth} and Tables~\ref{tab:lang} and~\ref{tab:projects} show that the curated corpus contains examples from multiple projects, monitor types, and implementation ecosystems.
The largest individual project contributes four cases, while the remaining cases span policy analyzers, output rails, proxy guardrails, verifiers, tool hooks, stream adapters, and model-client parsers.
Six cases occur outside Python, across TypeScript, Java, and C\#/.NET.
The CSV records 13 raw project labels, which we group into 11 framework project families; the established-boundary subset still spans eight project groups.

This breadth matrix matters because it separates family from framework.
The same project can expose multiple families at different safety integration points: request parsing, structured tool evidence, downstream policy consumers, or policy-state updates.
Conversely, the same family can appear in unrelated frameworks.
Carrier cases appear across policy analyzers, verifiers, stream callbacks, and breakpoint logic; adapter cases appear across output rails, proxy guardrails, and connector filters.
This descriptive pattern illustrates how the same dependency problem can arise at different safety/framework interfaces; it is not a population-level concentration result.

The project concentration table gives a second check on this interpretation.
LiteLLM contributes the largest number of cases, but those four cases exercise different boundaries: proxy guardrail input construction, stream-derived tool evidence, in-memory policy updates, and pass-through lifecycle hooks.
Promptfoo contributes verifier and tool-cap cases rather than output-rail cases; NeMo contributes both output mapping and multimodal prompt materialization.
This within-project heterogeneity shows that the curated set is not merely multiple instances of one subsystem-level repair.

The language split is also not the main claim, but it is a useful guardrail against overfitting the taxonomy to one implementation style.
Python projects dominate the available LLM safety ecosystem, so Python cases are expected.
The curated examples also include TypeScript, Java, and C\#/.NET implementations, where safety logic likewise consumes a structured representation whose shape or routing differs across framework boundaries.

\subsection{RQ3: How Far Do the Misses Reach?}

Breadth alone does not establish safety impact.
We next ask how far each confirmed repair-boundary failure reaches into the monitor stack: does it merely affect an observable trace field, or does it reach a blocking, routing, filtering, or verifier decision?
Table~\ref{tab:impact} separates E1 into operational forms alongside the E2 and E3 boundaries.

Table~\ref{tab:outcomes} is not a second taxonomy.
It summarizes the observable parent-to-fixed outcome shapes that appear across the families above.

Seventeen cases are Tier A confirmations over framework- or module-level boundaries; one isolates the changed logic.
Fourteen reach E1 enforcement: five blocking/filtering/rewriting surfaces, three interrupt or HITL routes, and six hook/cap/policy-state boundaries.
They span four families, ten framework project groups, and four implementation ecosystems, with seven native actions, six configured boundaries, and one synthetic-consumer consequence.
Thus E1 captures operational decisions---not merely internal trace differences---that the parent misses and the fixed version restores.
Appendix Table 11 gives the per-case 7/6/1 mapping.

Operationally, the blocking/filtering/rewriting cases permit or emit content that the fixed version stops or transforms; interrupt/HITL cases continue tool execution past a configured stop point; and hook/cap/policy-state cases fail to invoke, bind, or refresh an existing enforcement mechanism.
These distinct outcomes share one structure: the framework continues while the configured policy boundary misses the decision restored by the fixed revision.

E2 cases show verifier or safety-prompt inputs losing evidence that downstream evaluation relies on, while E3 captures structured evidence visibility at an observability boundary.
Reporting these strata separately keeps the operational action results distinct from evidence-channel results.

Table~\ref{tab:outcomes} summarizes the reproduced contrasts.
In the parent, evidence is absent, stranded, misrouted, or evaluated under stale policy state; in the fixed revision, the canary reaches the expected field and, where applicable, the intended action fires.

\begin{table}[t]
\centering
\caption{Before/after outcome patterns observed in the reproduced cases.}
\label{tab:outcomes}
\small
\begin{tabular}{llll}
\toprule
Outcome & Evidence & Decision & Repair \\
\midrule
Evidence invisible & absent & none & restore carrier \\
Enforcement skipped & visible & none & restore adapter \\
Wrong authority & stale & wrong & select channel \\
Policy state stale & stale & stale & rebind rule \\
\bottomrule
\end{tabular}
\end{table}

\subsection{RQ4: How Does the Protocol Validate Confirmed Cases?}

We evaluate the confirmation rule with candidate audits, negative controls, known-pattern fixtures, a runnable diagnostic comparison, and a frozen prospective pass.

\begin{table}[t]
\centering
\caption{Validation checks for the confirmation rule and canary-contract schema.}
\label{tab:validation}
\small
\begin{tabular}{p{0.28\linewidth}p{0.2\linewidth}p{0.2\linewidth}p{0.2\linewidth}}
\toprule
Validation check & Reviewed & Result & Contribution \\
\midrule
High-signal denominator & 195 refs & 9 newly promoted & candidate expansion \\
Random merged PR batch & 55 PRs & none confirmed & random-batch control \\
Held-out high-risk audit & 8 candidates & none confirmed & conservative promotion \\
Reviewed negatives & 13 candidates & excluded & boundary discipline \\
Known-pattern fixtures & 14 fixtures & 14 contracts & contract expressiveness \\
Rejected boundary cases & 3 cases & excluded & exclusion consistency \\
Mixed-candidate audit & 21 candidates & 18 / 3 split & rule stability \\
\bottomrule
\end{tabular}
\end{table}

The 195-reference audit expands the candidate pool, the mixed audit reproduces the 18/3 boundary split, and the negative controls exclude adjacent changes.
Known-pattern fixtures independently express all four family schemas with visibility and decision metadata.

Across five representative cases---Invariant, NeMo regex output mapping, CrewAI, Haystack, and LangChain stream redaction---ordinary functional checks pass without the canary oracle.
Visibility alone is insufficient for the NeMo and Haystack adapter/decision misses; partial oracles also fail to record the fixed repair.
The full contract combines changed-surface placement and parent/fixed visibility with decision assertions at action boundaries.
Appendix Table 10 reports the case-level comparison; the artifact includes its machine-readable records.

We also froze a follow-up audit rule before reviewing recent changes.
A first pass over 300 recent closed pull requests, 186 merged, produced 27 term-matched candidates.
Of these, 23 were excluded, three remained under reproduction, and one additional repair-boundary case was confirmed outside the retrospective 18-case corpus.

\section{Related Work}

Guardrails and agent safety.
Guardrail and red-teaming work evaluates detection once evidence reaches an expected interface \citep{wei2023jailbroken,mazeika2024harmbench,inan2023llamaguard,han2024wildguard}; agent studies expand that interface to actions, traces, tool calls, streams, artifacts, and provider schemas \citep{yao2023react,schick2023toolformer,qin2023toolllm,zhan2024injecagent,debenedetti2024agentdojo,andriushchenko2025agentharm}.
We instead ask whether a supported framework representation leaves evidence outside the monitor's view or strands a positive result before action, even while the path remains operational.
The distinction is operational: model-centric evaluation asks whether a model or detector recognizes a payload once it is presented through the chosen interface.
A framework-validity check asks whether the exercised representation reaches that interface and whether a positive result reaches the intended action.
The two can diverge: recognition can succeed while an adapter, stream normalizer, or authority choice prevents enforcement.

Runtime monitoring and software evolution.
Runtime verification studies enforcement, API evolution studies client breakage, and differential or metamorphic testing compares related executions \citep{leucker2009runtime,falcone2012runtime,dig2006api,rothermel1997regression,mcKeeman1998differential,chen2018metamorphic}; ML systems work documents hidden interface coupling \citep{sculley2015hidden,breck2017mltestscore,amershi2019se4ml}.
Canary contracts specialize these ideas to LLM safety with three constraints: a policy-intent anchor, payload placement only on the changed framework surface, and separate visibility and enforcement oracles.
Together they distinguish an evidence-path failure from an arbitrary revision difference.
Runtime verification supplies enforcement semantics, software-evolution testing supplies the parent/fixed contrast, and ML-systems work motivates scrutiny of hidden coupling.
Canary contracts join those pieces at the changed representation, making evidence placement and action routing explicit rather than treating every behavioral difference as equally relevant.

Our unit is the policy-linked evidence path through a framework boundary, enabling deterministic oracles across schemas, streams, adapters, authority channels, and policy-state updates.

\section{Conclusion}

Monitor-validity failure is a framework-interface failure: an operational path fails to carry evidence or a positive detector result to its policy boundary.
Across 18 confirmed cases and 11 framework project groups, the pattern recurs in carriers, adapters, authority channels, and new surfaces; 14 cases reach E1 enforcement boundaries.
Canary contracts expose it by combining changed-surface payload placement, visibility evidence, and action-boundary decision checks across parent and fixed revisions, turning hidden safety assumptions into explicit contracts for schemas, streams, tools, adapters, and policy-update paths in evolving LLM and agent infrastructure.

\bibliography{references}

\end{document}